\documentclass{article}

\usepackage[final]{neurips_2022}

\usepackage{graphicx}
\usepackage{color, colortbl}
\usepackage[utf8]{inputenc} % allow utf-8 input
\usepackage[T1]{fontenc}    % use 8-bit T1 fonts
\usepackage{hyperref}       % hyperlinks
\usepackage{url}            % simple URL typesetting
\usepackage{booktabs}       % professional-quality tables
\usepackage{amsfonts}       % blackboard math symbols
\usepackage{nicefrac}       % compact symbols for 1/2, etc.
\usepackage{microtype}      % microtypography
\usepackage{xcolor}         % colors
\usepackage{algorithmic}
\usepackage{graphicx}
\usepackage[linesnumbered,ruled]{algorithm2e}
\usepackage{amsmath}
\usepackage{empheq}
\usepackage{cleveref}
\usepackage{bm}

\crefname{section}{§}{§§}
\title{LogiGAN: Learning Logical Reasoning via Adversarial Pre-training}

\author{
    Xinyu Pi$^{1}$\thanks{indicates equal contribution. Work done during internship at Microsoft Research Asia.},~~Wanjun Zhong$^{2*}$,~~Yan Gao$^3$,~~Nan Duan$^3$,~~Jian-Guang Lou$^3$\\
    $^1$University of Illinois Urbana-Champaign, Urbana, USA 
    \\$^2$Sun Yat-Sen University~~~~$^3$Microsoft Research Asia \\
    \texttt{xinyupi2@illinois.edu},~~
    \texttt{zhongwj25@mail2.sysu.edu.cn}\\
    \texttt{\{yan.gao, jlou, nanduan\}@microsoft.com}\\
}

\begin{document}

\maketitle

\begin{abstract}
We present LogiGAN, an unsupervised adversarial pre-training framework for improving logical reasoning abilities of language models.
Upon automatic identification of logical reasoning phenomena in massive text corpus via detection heuristics, we train language models to predict the masked-out logical statements.
Inspired by the facilitation effect of reflective thinking in human learning, we analogically simulate the learning-thinking process with an adversarial Generator-Verifier architecture to assist logic learning.
LogiGAN implements a novel sequential GAN approach that  
\textit{(a)} circumvents the non-differentiable challenge of the sequential GAN by leveraging the Generator as a sentence-level generative likelihood scorer with a learning objective of reaching scoring consensus with the Verifier;  
\textit{(b)} is computationally feasible for large-scale pre-training with longer target length. 
Both base and large size language models pre-trained with LogiGAN demonstrate obvious performance improvement on 12 datasets requiring general reasoning abilities, revealing the fundamental role of logic in broad reasoning, as well as the effectiveness of LogiGAN.
Ablation studies on LogiGAN components reveal the relative orthogonality between linguistic and logic abilities and suggest that reflective thinking's facilitation effect might also generalize to machine learning \footnote{The code is released in \url{https://github.com/microsoft/ContextualSP/tree/master/logigan}}.
\end{abstract}

% To tackle with both \textit{(a)} the non-differentiable challenge where 
% the gradient propagation from the Verifier side to the Generator is hindered by the discrete intermediate beam-search for sequential text generation, and \textit{(b)} the computational efficiency challenge to make adversarial training scalable to large-scale pre-training with long target sequence,
% LogiGAN implements a novel GAN approach for text generation:
% Instead of attempting to obtain high Verifier score with its generated examples, our Generator scores candidate statements and aims at reaching ranking consensus with the Verifier.

\section{Introduction} 

\begin{center} ``\textit{Learning without thinking is labor lost; thinking without learning is perilous.'' ~~ -- ~Confucius} \end{center}

% Transformer architectures, powered by the attention mechanism \citep{NIPS2017_3f5ee243}, is springing up a new tide of artificial intelligence, bringing radical revolution to almost all sub-communities.
% As a primary beneficiary, Pre-trained Language Models (PLMs) from the Natural Language Processing (NLP) community leverages the information expressiveness advantage of transformer architectures, becoming increasingly more competent at capturing diminutive contextualized linguistic variations.
% A great line of pre-training research \citep{DBLP:journals/corr/abs-1810-04805, DBLP:journals/corr/abs-2005-14165, 2020t5},  is gradually conquers the complex nature of natural language, and PLMs are approaching human-level in numerous tasks requiring basic linguistic abilities \citep{DBLP:journals/corr/abs-1804-07461, DBLP:journals/corr/RajpurkarZLL16}.
% % requiring sophisticated understanding of human language.

% Despite the emerging fervor, researchers soon realize that there is a bottleneck that is insurmountable via merely improving linguistic abilities: PLMs are relatively incompetent in \textbf{reasoning} \citep{helwe2021reasoning, DBLP:journals/corr/abs-1911-03343}. Following this, the community's interest on reasoning reaches a new climax, with multitudinous aspects of concentration and various schools of interpretation.

Pre-trained Language Models (PLMs) \citep{DBLP:journals/corr/abs-1810-04805, DBLP:journals/corr/abs-2005-14165, 2020t5} are approaching human-level performance in numerous tasks requiring basic linguistic abilities \citep{DBLP:journals/corr/RajpurkarZLL16,DBLP:journals/corr/abs-1804-07461},
setting off a huge wave of interest in Natural Language Processing (NLP).
Despite the emerging fervor, researchers soon realized that PLMs are relatively incompetent in their \textbf{reasoning} abilities, which seems to be an insurmountable bottleneck for PLMs with even better linguistic abilities \citep{DBLP:journals/corr/abs-1911-03343,helwe2021reasoning}. 
Following this, researchers delve into reasoning from multitudinous aspects, striving to improve PLMs' reasoning abilities.

From our perspective, reasoning (in natural language) is essentially an inferential process where an unstated statement is drawn based on several presented statements, and \textbf{Logic} is the systemic set of principles that provides reasoning with correctness and consistency assurance \citep{Hurley1982-HURACI}.
Regardless of the variability of contents, logical reasoning generally incorporates two invariant forms: drawing conclusions based on some premises (aka. deduction \& induction, \citep{reichertz2013induction}), or hypothesizing premises to explain some conclusions (aka. abduction \citep{sep-abduction}).
Most existing tasks requiring general reasoning ability, such as natural language inference \citep{DBLP:journals/corr/abs-1910-14599} and complex machine reading comprehension \citep{lai2017large}, can be readily interpreted by this criterion. 
Other tasks requiring specialized reasoning skills can be considered either as 
(i) providing sufficient premises but requiring specific ways of premise extraction to draw conclusions, such as multi-hop \citep{yang2018hotpotqa} or hybrid \citep{chen-etal-2020-hybridqa} reasoning; 
or (ii) requires external knowledge, such as commonsense \citep{sap-etal-2020-commonsense} or numerical \citep{Dua2019DROPAR} knowledge, as premises to draw conclusions, hence could also be interpreted by the two forms of logical reasoning. 
Following this analysis on the relation between logic and reasoning, \textit{Logic ability} will be an essential foundation for a broad scope of reasoning, and should be prioritized in improving PLMs' reasoning abilities
\footnote{We expand this analysis in-depth in App. A, and refer intrigued readers there.}.

% However, discovering or invoking relevant premises is one thing, to produce proper conclusions based on such premises (i.e., reasoning logically) is another.
% An abundance of neural-symbolic reasoning models, which first leverages PLMs to parse inputs into intermediate human-defined symbolic representations (i.e., generate premises), then outsources leftover reasoning steps to rule-based executors or specialized neural modules to assist production of final results.
% This sheds light on PLMs' intrinsic lack of ability in logical reasoning. 

Conventional pre-training via \textit{randomized} \textbf{M}asked \textbf{L}anguage \textbf{M}odeling (MLM) and auxiliary tasks are generally developed upon \cite{Firth1957}'s distributional hypothesis of semantics -- "a word is characterized by the company it keeps."
Under this paradigm, models efficiently learn to capture grammatical structures and contextualized semantics.
However, since logical consistency is beyond correctness on a linguistic level, it is less obvious how MLM could help with logical reasoning abilities.
Do models harvest logic ability for free from MLM?
Or is that something that needs further learning beyond language acquisition?
Motivated by these questions, we propose an \textbf{unsupervised pre-training method aiming at enhancing the logical reasoning ability of PLMs}:
we automatically identify occurrences of logical reasoning phenomena in large corpus via detection heuristics, and then require PLMs to predict the masked-out logical statements made in the original context (Section \ref{sec:logic-pretrain}).
% For example, in the case ``Bob recently got obsessed with junk food. Today he went on the scale and was shocked to observe his huge weight gain. He makes up his mind to lose weight. \textit{Therefore}, \texttt{[MASK]}'', the prediction goal is masked original text ``he decides to go on a diet''.
For example, in the case ``Bob recently made up his mind to lose weight. \textit{Therefore}, \texttt{[MASK]}'', the prediction goal is the masked original statement ``he decides to go on a diet''.

However, statements different from the original statement could also be logically consistent with respect to a given context.
For example, ``he decides to exercise from today on'' is also a reasonable inference in the case above. 
% Due to the diverse nature of logical consistent statements, learning only from a single ground-truth statement make Generator not to be encouraged in 
Since Generators trained merely from recovering original statements are not encouraged to explore the possibilities of other reasonable statements, their overall learning effectiveness of logic could potentially be degraded.
Therefore, it makes sense to provide additional feedback based on the degree of logical consistency between statements predicted beforehand and the original context -- we realize this much resembles humans' reflective thinking process.
Inspired by research from cognitive psychology \citep{moon2013reflection, boud2013reflection, di2016making} advocating for the vital role of reflective thinking in improving the experiential efficiency of human learning,
we hypothesize that machines might also benefit from reflective thinking in their learning processes.
Following this hypothesis, we analogically simulate humans' learning-thinking process with a Generator-Verifier architecture, 
% similar to \citep{cobbe2021gsm8k, DBLP:journals/corr/abs-2201-11903, DBLP:journals/corr/abs-2003-10555, nijkamp-etal-2021-script}, 
and propose \textbf{LogiGAN}, a novel adversarial training approach tailored for sequential GAN training to further facilitate the learning of logical reasoning.

% However, unlike the MLM training scenario where gold answers always fall into a pre-defined massive vocabulary list, the gold answer space of logical reasoning is infinite.
% This intrinsic sparsity nature of logical reasoning calls for better data efficiency (i.e., drawing more information from a single training example) under limitations of computation power.

In LogiGAN's design, the Generator learns not only to recover the original masked statements, but also to score candidate statements (based on their generative likelihood) in a manner that could reach scoring consensus with the Verifier, who learns to make judgments on the logical consistency between premises and conclusions.
The more logically consistent the Verifier thinks of a statement w.r.t. the input context, the higher generative likelihood score is expected to be assigned by the Generator.
% Combined with diversified sampling strategy for candidate statements generation, the adversarial mechanism encourages the Generator to explore broader possibilities of reasonable statements other than the original one.
To encourage the exploration of broader possibilities of reasonable statements other than the original one, we also apply a diversified sampling strategy for candidate statement generation.
Both Generator and Verifier scoring processes are continuous throughout the adversarial training, thereby circumventing the non-differentiable barrier in sequential GAN posed by the discrete beam-search.
Moreover, LogiGAN does not involve token-wise Monte Carlo Search for policy gradient estimation, and scoring processes of Generator and Verifier are decoupled, so that parallel score computation is possible.
This makes large-scale pre-training with longer target length computationally feasible.

To test the effectiveness of LogiGAN, we extensively experiment on $\mathbf{12}$ datasets requiring general reasoning ability.
The apparent performance improvements of \textit{both base and large size PLMs} across all tasks reveal models' harvest of logic ability, shoring up the fundamental role of logic in general reasoning.
We also carry out ablation studies to understand the functionality of LogiGAN components, the results of which shed light on the relative orthogonality between linguistic and logic ability and suggest that the facilitation effect of reflective thinking is also generalizable to machine learning.

\section{Logic Pre-training}
\label{sec:logic-pretrain}

In this work, we primarily focus on improving PLMs' ability of \textit{informal logic} \citep{sep-logic-informal}.
We include the three most classical types of logical reasoning: \textbf{deductive, inductive, and abductive reasoning} conducted in the form of natural language \citep{reichertz20044, kennedy2018deduction, reichertz2007abduction, sep-abduction} in our consideration.
Note that our coverage is broader than the informal logic strictly defined in the philosophy community \citep{Munson1976-MUNTWO-3} that primarily focuses on analyzing the soundness and cogency of the application of the aforementioned reasoning in real-life arguments.
The other half of logic investigation  -- the normative study of formal logic (typically conducted in a symbolic form), such as truth-function logic \citep{buvac1993propositional}, modal logic \citep{priest2008introduction}, and fuzzy logic \citep{483332}, is beyond the scope of this paper.

% \textbf{Verbal reasoning} \citep{hitch1976verbal}, the reasoning that emphasizes less on whether a set of reason \textit{factually entails} a conclusive statement but more on sentential relations and concept framing in words, also falls into the scope of our work.
% For example, in the restaurant review, ``Their environment is noisy. The food, however, \texttt{[MASK]}'', people effortlessly know some statements positive polarity is coming up.
% This reasoning is done on basis of polarity understanding of the word ``noisy", as well as the contrast function of adverb ``however'' (i.e., reasons over language to derive harmonic semantics), and is therefore distinct from the informal logic that reasons over facts to derive the truth.
% The abundance of verbal reasoning in the US's most widely taken graduate examinations (GRE, GMAT, and LSAT) reveals its importance and shores it up as a relative independent type of logic from informal logic.

% \subsection{Logic Indicators as Philosophical Heuristics}
\textbf{Logic Indicators as Detection Heuristics.}
To set up an unsupervised pre-training aiming at improving models' logic ability, the very first step will be to identify logical reasoning phenomena from a vast-scale unstructured text corpus.
While invocations of logic are not explicitly stated in most cases, writers' usage of \textit{logic indicators} usually marks their logical reasoning processes with high precision \citep{Hurley1982-HURACI}, thereby serving as an ideal heuristic device for our detection purpose.
We consider two standard types of logic indicators:
\textbf{(i) \textit{conclusion indicator}} such as ``Therefore'', ``We may infer that'', which denotes drawing conclusion deductively or inductively from given premises;
And \textbf{\textit{(ii) premise indicator}} such as ``Due to'', ``The reason that'', which denotes abductively hypothesizing premises that explain or provide evidence to some stated conclusions.

\begin{figure*}[t]
    \centering
    \includegraphics[width=\textwidth]{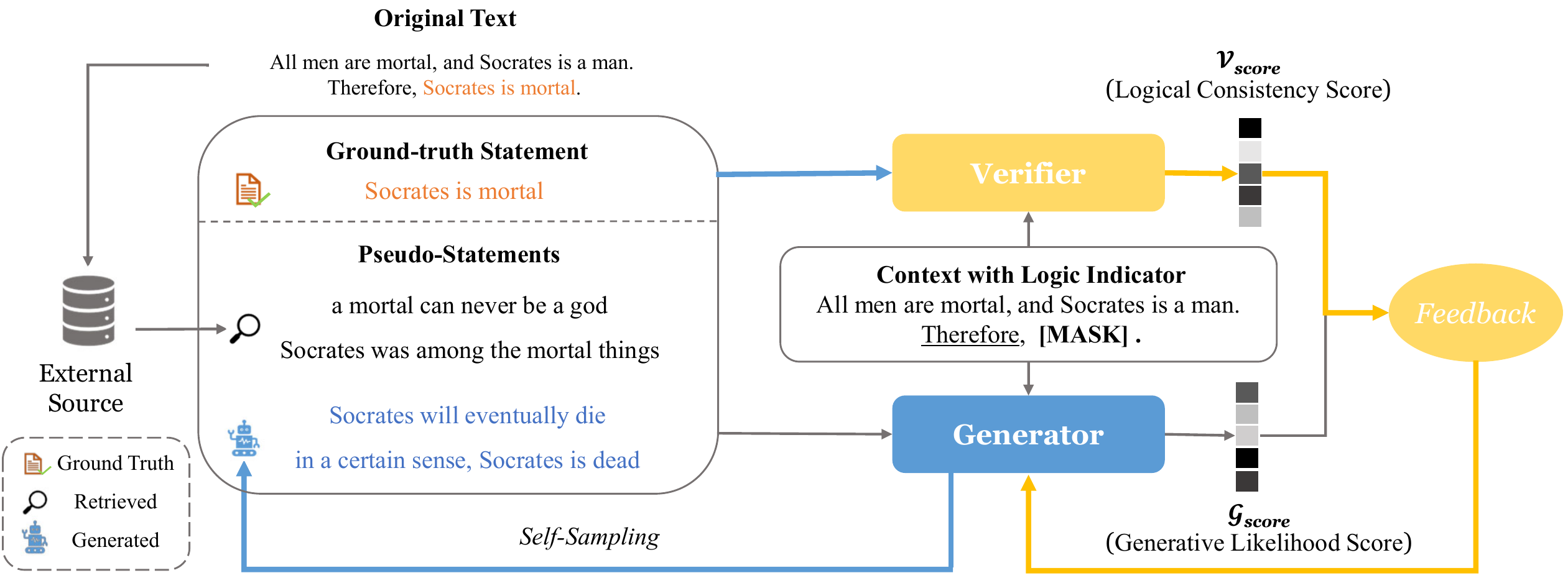}
    \caption{LogiGAN Overview. Generator targets to predict the masked-out logical statement and scores candidate statements, while Verifier justifies the logical correctness of statements. The blue path indicates the process where the Generator helps Verifier learning, while the yellow path denotes the process of giving Verifier feedback for Generator training.}
    \label{fig:overview}
    \vspace{-6mm}
\end{figure*}

% \subsection{Corpus Construction} \label{sec:corpus}
\textbf{Corpus Construction.} \label{sec:corpus}
For a text corpus, we detect every occurrence of pre-defined logic indicators (listed in App. C), and mask out (i.e., replace with \texttt{[MASK]}) the entire \textbf{statement} subsequent to the indicator (each training example will have exactly one masked-out statement).
Then models' task will be to perform language modeling and predict the masked statement.
We emphasize that \textbf{statements} are declarative sentences or declarative clauses, owning complete subject and predicate structures, and are capable of being factually true or false.
To supply sufficient context information for these predictions, we keep $x$ complete sentences previous to the \texttt{[MASK]}, as well as $y$ sentences after the \texttt{[MASK]}, where $x$ and $y$ can be sampled from a geometric distribution with pre-defined hyper-parameters.
Fig. \ref{fig:overview} illustrates the input and output format, and we discuss details in Sec. \ref{sec:setup}.

% \subsection{Masked Statement Prediction}
\textbf{Masked Logical Statement Prediction.~}
In the simplest setting, the Generator learns to infill the masked statement via a \textit{single-task} pre-training, which fulfills the training process of a typical masked language modeling task.
The only difference is that models no longer predict \textit{\textbf{randomly masked tokens or spans}} but instead \textit{\textbf{logic-targeted masked complete statements}}.
Models are trained to perform Max Likelihood Estimation (MLE) for masked statements under a typical teacher forcing loss.
Practically, generative pre-trained language models such as
T5 \citep{2020t5}
% BART \citep{DBLP:journals/corr/abs-1910-13461}, , and GPT \citep{DBLP:journals/corr/abs-2005-14165}
could take up the position of Generator $\mathcal{G}$.
Given a single input context / output statement pair $(c, s)$, the teacher forcing loss can be mathematically expressed as \footnote{$w_t(.)$ denotes the $t^{th}$ token of a input string.}:
\begin{equation}
\small
\mathcal{L}_{tf}(c, s) = - \frac{1}{T}\sum_{t=1}^{T}\log p_{\mathcal{G}_\theta} \left(w_t(s) ~|~ w_{1:t-1}(s); c \right)
\label{equ:teacher-forcing}
\end{equation} 

\section{The Adversarial Training Framework} \label{sec:logigan}
\label{sec:logic-pretrain}
% \begin{figure*}[t]
%     \centering
%     \includegraphics[width=\textwidth]{Figures/logiGAN.pdf}
%     \caption{LogiGAN Overview. Generator targets to predict the masked-out logical statement and scores candidate statements, while Verifier justifies the logical correctness of statements. The blue path indicates the process where the Generator helps Verifier learning, while the yellow path denotes the process of giving Verifier feedback for Generator training.}
%     \label{fig:overview}
%     % \vspace{-3mm}
% \end{figure*}
% This section introduces the design details of the adversarial training of LogiGAN.
Since Generators trained merely from recovering masked original statements miss out opportunities of exploring other reasonable statements, LogiGAN implements an adversarial mechanism for providing Generators with extra signals based on logical consistency between pseudo-statements (sampled from Eq. \ref{equ:3}) and context to encourage explorations.
The adversarial framework has two major components:
(i) a \textit{Verifier} $\mathcal{V}$ that learns to judge logical consistency between statements and context;
(ii) a \textit{Generator} $\mathcal{G}$ that learns both to recover masked original statements, and scores pseudo-statements (based on their generative likelihood) in a manner that could reach scoring consensus with the Verifier -- The more logically consistent the Verifier thinks of a statement w.r.t. the input context, the more likely the Generator is expected to generate the statement under the input context (i.e., assign higher generative likelihood score).
The overall objective of LogiGAN can be expressed as the minimax objective:
\begin{equation} \label{equ:2}
\small
J^{\mathcal{G}^*.\mathcal{V}^*} = \min_{\mathcal{\theta}}\max_{\mathcal{\phi}} ~\mathbb{E}_{\bm{s}^+ \sim p_{\text{true}}(.|c)} [\log \mathcal{V}_{\phi}(c, \bm{s}^+)] + \mathbb{E}_{\bm{s^ -} \sim  p_{\text{neg}}(. | \mathcal{G}_\theta, c, s^+)} [\log (1 - \mathcal{V}_{\phi}(c, s^{-}))]. 
\end{equation}
where $\mathcal{G}_\theta$ and $\mathcal{V}_{\phi}$ denote  the Generator and the Verifier with model parameters $\theta$ and $\phi$, respectively. $s^+/s^-$ represents ground-truth statements from original text / sampled pseudo-statement
\footnote{Note: in real practice, there is a \textbf{gap} between sampled \textit{pseudo-statements} $s^-$ and \textit{logically inconsistent} statements.  We keep current symbolic denotations for simplicity only and discuss this issue in App. B.}.
We discuss sampling details of pseudo-statements later in this section in Eq. \ref{equ:3}.

Classical GAN settings \citep{https://doi.org/10.48550/arxiv.1406.2661, DBLP:journals/corr/ZhuPIE17} fall short in sequential generation because the gradient propagation from the Verifier to the Generator is blocked by a non-differentiable beam-search during text generation. 
Previous approaches such as \citep{10.5555/3298483.3298649} address this challenge by token-wise policy gradient estimation via Monte Carlo Search.
However, since the sampling run-time grows exponentially with the length of the target sequence, their original implementations are not applicable to million-scale pre-training with relatively longer target length as in our scenario.

% instead circumvents the non-differentiable process by additionally leveraging the Generator as a scorer to assign continuous scores to statements, thereby enabling back-propagation.
% and could scale to arbitralily long output sequences, and has higher efficiency to be applied in pre-training.
% GAN-loss of our design could scale to arbitrarily long output sequences.
% Our special design of GAN-loss circumvents the two most prickly issues in applying GAN-alike training on language models:
 
Different from them, LogiGAN omits the token-wise Monte Carlo Search for policy gradient estimation, and realizes a similar goal via measuring the similarity of scoring distributions between Verifier and Generator.
% Different from them, LogiGAN omits the sequential Monte Carlo Search for token-wise policy gradient estimation, and instead estimate an overall gradient with sentence-level scoring consensus between Generator and Verifier.
The main procedures of LogiGAN can be summarized in four steps: 
\textit{(a)} several candidate pseudo-statements are sampled on a sentence level;
\textit{(b)} the Verifier assigns the \textit{\textbf{logical consistency scores}} $\mathcal{V}_{score}$ based on how logical consistent these candidates are w.r.t the original context; 
\textit{(c)} the Generator assigns the sentence-level \textit{\textbf{generative likelihood score}} $\mathcal{G}_{score}$ to each candidate to reflect how likely it will produce the pseudo-statement under the given context.
\textit{(d)} The similarity between Generator and Verifier score distributions is computed as a new signal to encourage the Generator to reach scoring consensus with the Verifier -- i.e., the more logically consistent the Verifier thinks of the statement, the higher likelihood score the Generator is expected to assign.
Since both scoring processes are continuous, the non-differentiable barrier is successfully bypassed.
Meanwhile, this design does not involve sequential token-level sampling and decouples the Generator and Verifier scoring processes, thereby enabling parallel score computations.
This makes large-scale pre-training with relatively longer target sequence length computationally feasible.

% Meanwhile, this design avoids token-level Monte Carlo sampling, and enables parallel score-computation has high computational efficiency, because the decoupled score-computation process can be done parallel for Generator and Verifier, making large-scale adversarial pre-training with arbitrarily long target sequence feasible. 
% Finally, we expect the statement with higher $\mathcal{G}_{\text{score}}$ (more similar to the generated sequence of $\mathcal{G}$), will also receive higher $\mathcal{V}_{\text{score}}$.
% Afterwards, the differentiable score-computation process enables feedback from Verifier to propagate to the Generator.

% Then Verifier (\cref{sec:Verifier-training}) is trained to discriminate the pseudo-statements, while the Generator is trained both to generate logical consistent statements and to reach scoring consensus with the Verifier.

The overall framework overview is illustrated in Fig. \ref{fig:overview}, and the detailed training procedure is summarized in Algorithm \ref{alg:training}.
To diversify the candidate pseudo-statements, we sample pseudo-statements from two sources:
(i) self-sampling via diversified beam-search from the Generator;
or (ii) retrieving similar statements from the corpus, and the sampling process can be summarized as:
\begin{equation} \label{equ:3}
p_{\text{neg}}(. ~|~ \mathcal{G}_\theta, c, s^+) = \{s_\alpha \sim \mathcal{G}_\theta{(. ~|~ c)} ~~\cup~~ s_{{\beta}} \sim R(s^+)\},
\end{equation}
where $\mathcal{G}_\theta{(. ~|~ c)}$ denotes self-sampled statement $s_\alpha$ given context $c$ from Generator $\mathcal{G}_\theta$, and $R(s^+)$ denotes a retriever\footnote{Any retriever is feasible and we adopt BM25 as the retrieving method here.} that retrieves textually similar statements $s_{\beta}$ with ground-truth statement $s^+$ from the corpus. 
Note that this process is conducted separately for the corpus of Verifier and Generator. 

\begin{algorithm}[h!]

\caption{Adversarial Training Process} 
\label{alg:training} 
% \small
\SetKwInOut{Input}{Dependencies}
\SetKwInOut{Output}{Output}
\Input{(1) A Pre-trained Generative Language Model as Generator $\mathcal{G}_0$\\
(2) A Pre-trained Discriminative Language Model as Verifier $\mathcal{V}_0$\\ 
(3) Generator Source Training Coprus $C_\mathcal{G}$ with size $M$\\
(4) Verifier Source Training Corpus $C_\mathcal{V}$ with size $N$\\
(5) Pre-defined Warmup epochs $E$, max iterations of GAN training $Q$\\
(6) Pre-defined training sample size $m$, $n$ for $\mathcal{V}$, $\mathcal{G}$ per iteration\\}

Random partition $C_\mathcal{G}$ into $C_{\mathcal{G}_\alpha}$, $C_{\mathcal{G}_\beta}$ with size $M_\alpha$, $M_\beta$;\\
$\mathcal{G}_0$ $\leftarrow$ Warmup $\mathcal{G}_\alpha$ on $C_{\mathcal{G}0}$ for $E~epochs$ with $\mathcal{L}_{tf}$;\\
\For{~i ~in~  1:Q ~}{
    $\mathcal{G}_{i} \leftarrow \mathcal{G}_{i-1}$;\\
    ${{C_\mathcal{V}}_i, C_\mathcal{G}}_i \leftarrow$ Sample $m$ examples from $C_\mathcal{V}$, and $n$ examples from $C_\mathcal{G_\beta}$, w/o replacement; \\
    $\widetilde{{C_\mathcal{V}}_i}, \widetilde{{C_\mathcal{G}}_i} \leftarrow$  Sample pseudo-statements for ${C_\mathcal{V}}_i$, ${C_\mathcal{G}}_i$ with  $\mathcal{G}_{i}$ and BM25, as in Eq. \ref{equ:3};  \\
    $\mathcal{V}_i \leftarrow$ Train  $\mathcal{V}_{i-1}$ on $~\widetilde{{C_\mathcal{V}}_i}~$ for $1~epoch$ with $\mathcal{L}_{ver}$, as in Eq. \ref{equ:4};  \textcolor{orange}{~~(Verifier Training)} \\
    \For{~$\widetilde{c}$  ~in~ batch~(~$\widetilde{{C_\mathcal{G}}_i}$~)~~}{
        $\mathcal{V}_{score}, \mathcal{G}_{score} \leftarrow$  $\mathcal{V}_{i},~ \mathcal{G}_{i}~$  do scoring on $~\widetilde{c}$, as in Eq. \ref{equ:5} and \ref{equ:6}; \\
        $\mathcal{L}_{gen} \leftarrow  \lambda_1~\mathcal{L}_{tf}(s^+\ \text{from}~ \widetilde{c}) +  \lambda_2~D_{KL}(\mathcal{V}_{score} ~||~ \mathcal{G}_{score})$ , as in Eq. \ref{equ:7};  \\
        $\mathcal{G}_{i} \leftarrow$ Update $~\mathcal{G}_{i}~$ for $1~step$ with $\mathcal{L}_{gen}$; \textcolor{cyan}{~~(Generator Training)}
    }
    % \EndFor
}
% \EndFor
\end{algorithm}
% \vspace{-6mm}

\subsection{Training of Verifier} \label{sec:Verifier-training}

The Verifier serves as a critic to judge whether a statement is logically consistent w.r.t. the context. 
Therefore, the training task of Verifier can be viewed as a binary classification problem. 
Pre-trained language models that could perform discriminative classification tasks such as BERT \citep{DBLP:journals/corr/abs-1810-04805}, ALBERT \citep{DBLP:journals/corr/abs-1909-11942}, and RoBERTa \citep{DBLP:journals/corr/abs-1907-11692}, will be suitable for the role of Verifier. 
With $y=1$ for both ground-truth and logically consistent pseudo-statements, and $y=0$ for other pseudo-statements, 
the binary classification loss for a single pair of context/statement/label $(c, s, y)$ of Verifier can be mathematically expressed as (omitting average):
\begin{equation} \label{equ:4}
    \mathcal{L}_{ver}(c, s, y) = -y \log \mathcal{V}_\phi(y ~|~ [c; s]) - (1 - y) \log (1 - \mathcal{V}_\phi(y ~|~ [c; s])),
\end{equation}
% When training the Verifier, the positive statement is the ground-truth statement, whereas the negative statements consist of both the retrieved statements and the statements produced by the Generator. 
% Thus, the improvement of generative ability of the Generator will also enhance Verifier training.

\subsection{Training of Generator}   \label{sec:analogy}

% Here we convey the fundamental intuition of our design of GAN-loss by drawing an analogy: Suppose there is an $100\%$ honest poet (i.e., a generative model) called G, who is well-trained for composing but not criticizing poems, and we wish to measure G's skills of poetry (i.e., ability of generating desirable content).
% To serve this purpose, we sample various pre-defined themes or subject-matters (i.e., various textual context in corpus) and hope to see G's behavior under these themes.
% Instead of requiring G to actually write a large batch of works from scratch and evaluate them one-by-one, we could instead show G some of existing works that is likely to fit into the given theme, either composed by G or by others poet, and demands G to \textbf{rank} the displayed works based on the his/her likelihood of producing something qualitatively and stylishly similar (i.e., compute likelihood).
% The ranking provided by G is then checked against the ranking from the other authoritative art critic (i.e., Verifier score), the agreeableness of which will then serve an objective evaluation of G's poetry skills could be formed -- If G is good at poetry, then G should also be able to properly appreciate existing works.

The Generator targets both to recover the original masked statements, and to score pseudo-statements based on their generative likelihood in a manner that could reach sentence-level scoring consensus with the Verifier.
This corresponds to the two sources of learning signals received by the Generator:
(i) the original generative objective with teacher forcing loss defined in Eq.\ref{equ:teacher-forcing} as a signal; and
(ii) the distribution similarity between sentence-level generative likelihood score assigned by Generator and logic consistency score assigned by Verifier.
To achieve the goal of (ii), we first sample pseudo-statements $\{s^{-}_1, ..., s^{-}_{n}\}$ from $p_{\text{neg}}(. ~|~ \mathcal{G}_\theta, c, s^+)$. Then the Verifier assigns \textbf{\textit{logical consistency score}} $\mathcal{V}_{\text{score}}$ based on how logically consistent the pseudo-statements are w.r.t. the context, expressed as:
\begin{equation} \label{equ:5}
\mathcal{V}_{\text{score}}(c; s^{-}_1, ..., s^{-}_{n}) = [ \mathcal{V}_{\phi} (s^{-}_1, c);~ \mathcal{V}_{\phi} (s^{-}_2, c);~ ...;~ \mathcal{V}_{\phi} (s^{-}_n, c)],
\end{equation}
After this, the Generator assigns a sentence-level \textbf{\textit{generative likelihood score}} $\mathcal{G}_{\text{score}}$ for each pseudo-statement to reflect how likely the pseudo-statement will be produced  under the given context:
\begin{equation} \label{equ:6}
\mathcal{G}_{\text{score}}(c; s^{-}_1, ..., s^{-}_{n}) = [ \ell_{\theta} (s^{-}_1 ~|~ c);~ \ell_{\theta} (s^{-}_2 ~|~ c);~ ...;~ \ell_{\theta} (s^{-}_n ~|~ c)],
\end{equation}
where $\ell_{\theta} (s ~|~ c)$ is the accumulated log-likelihood of the statement $s$ conditioned on the context $c$.

Afterward, each statement with a high Verifier score $\mathcal{V}_{\phi}  (s, c)$ is also expected to receive a high generative score $\ell_{\theta} (s ~|~ c)$  to facilitate the Generator's capturing of the Verifier's judgment criterion based on logic consistency. 
KL-divergence \citep{kullback1951information} $D_{KL}$ is therefore a appropriate measure for the similarity between the score distribution of $\mathcal{V}_{\text{score}}$ and $\mathcal{G}_{\text{score}}$.
For the purpose of smoothing the gradient to stabilize the GAN training process, we gather both the ground-truth (learned with teacher-forcing loss) and pseudo statements (learned with KL loss) inside the same batch w.r.t. a single input context $c$.
In our case, there is exactly one ground-truth statement and $n$ pseudo-statements for each input context $c$.
For a batch of $(c; s^+, s^{-}_1, ..., s^{-}_{n})$, the overall objective of the Generator is defined as (in App. F we show how Eq. \ref{equ:7} commits to the optimization of Eq. \ref{equ:2}): 
\begin{equation} \label{equ:7}
\mathcal{L}_{gen} (c; s^+, s^{-}_1, ..., s^{-}_{n}) = \lambda_1 ~ \mathcal{L}_{tf}(c, s^+) + \lambda_2 ~ D_{KL} (\mathcal{V}_{\text{score}}(s^{-}_1, ..., s^{-}_{n})) ~||~ \mathcal{G}_{\text{score}}(s^{-}_1, ..., s^{-}_{n}).
\end{equation}

% \begin{equation}
% \small
% J^{\mathcal{G}^*.\mathcal{V}^*} = \min_{\mathcal{\theta}}\max_{\mathcal{\phi}} ~\mathbb{E}_{\bm{s}^+ \sim p_{\text{true}}(.|c)}[\log \mathcal{V}_{\phi}(c, \bm{s}^+)] + \mathbb{E}_{\bm{s^ -} \in {\{s_\alpha \sim \mathcal{G}_\theta(. | c) ~\cup~  ES(s^+) \}}}  [\log (1 - \mathcal{V}_{\phi}(c, s^{-}))].
% \end{equation}

\section{Experiment Setup} \label{sec:setup}
\subsection{Datasets}
To test the effectiveness of LogiGAN,
we extensively experiment on $\mathbf{12}$ datasets requiring reasoning via natural language.
Specifically, ReClor \citep{yu2020reclor}, LogiQA \citep{10.5555/3491440.3491941}, Adversarial NLI - ANLI, \citep{DBLP:journals/corr/abs-1910-14599}, focuses especially on logical reasoning, 
TellMeWhy \citep{DBLP:journals/corr/abs-2106-06132} on abuductive reasoning,
HotpotQA \citep{DBLP:journals/corr/abs-1809-09600} on multi-hop reasoning,
QuoRef \citep{Dasigi2019QuorefAR} on reasoing with co-reference resolution, 
MuTual \citep{DBLP:journals/corr/abs-2004-04494}, DREAM \citep{sundream2018}), SAMSum \citep{gliwa-etal-2019-samsum} on reasoning in conversational scenarios, 
and NarrativeQA \citep{narrativeqa}, RACE \citep{lai2017large},
XSum \citep{xsum-emnlp} on general verbal reasoning.
These datasets make most, if not all, necessary premises for drawing logically consistent conclusions available in their provided context, and require few external premises like commonsense or numerical knowledge.
Hence, they fit nicely for testing our hypothesis that LogiGAN brings PLMs logic ability beyond their intrinsic linguistic ability, which could benefit general reasoning processes.

% Hence, they fit nicely for testing our hypothesis that the logic ability brought by LogiGAN could potentially benefit general reasoning processes beyond PLMs' intrinsic linguistic ability.

\subsection{Pre-training Corpus} 
\begin{figure*}[t]
    \centering
    \includegraphics[width=0.85\textwidth]{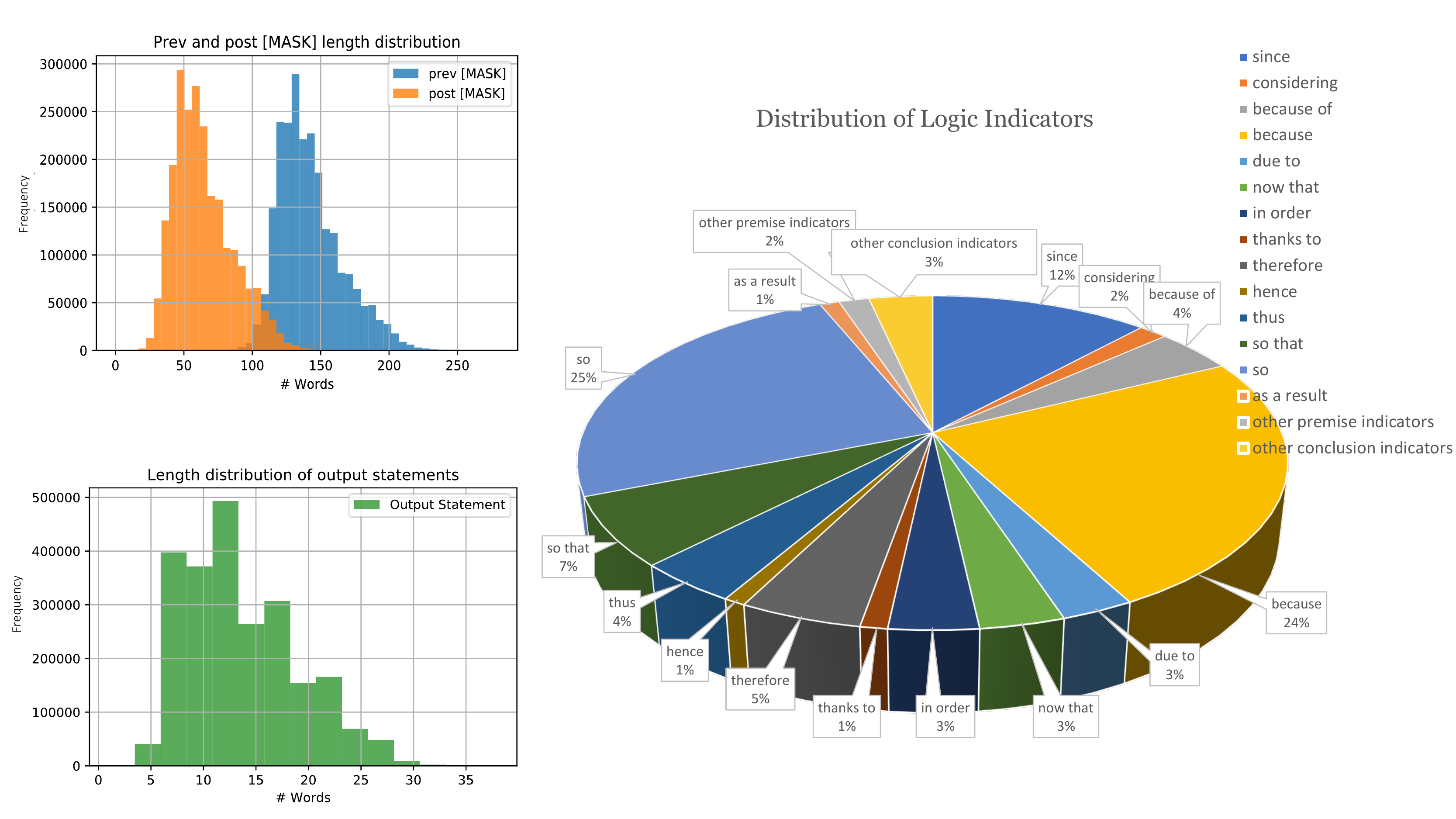}
    \caption{Corpus statistics. Histograms on the left side display length of masked statements (bottom) and prev-and-post statement context (top). The right-side pie chart displays indicators' distribution.}
    \label{fig:stat}
    \vspace{-4mm}
\end{figure*}
% \textbf{Corpus Construction Details}~~
We apply the corpus construction methodology (\cref{sec:corpus}) on the widely used \textit{BookCorpus} \citep{soskkobayashi2018bookcorpus}, which consists of e-books and movies with topics crawled from general domains.
Although some corpus featuring debates and arguments \citep{walker-etal-2012-corpus, Abbott2016InternetAC, swanson-etal-2015-argument} appears to be more suitable for our emphasis on logic, we do not elect them due to their high domain specificity in fields such as politics, law, and economics.
We discard overly short statements and instances where indicators do not indicate logical reasoning (e.g., ``\textit{since} 2010'' indicating a time point rather than premises,  ``\textit{so} happy'' indicating degree of the subsequent adjective rather than conclusions).
This results in $3.14$ million ($1.43$ and $1.71$ million from conclusion and premise indicators, respectively) instances. Corpus statistics are visualized in Fig. \ref{fig:stat}.

\subsection{Models}
\paragraph{Baseline Choice.} 
Since our primary goal of the experiment is to test the effectiveness of LogiGAN and test our hypothesis that logic ability can be further enhanced beyond PLMs' intrinsic linguistic ability, we only compare models pre-trained with LogiGAN against their vanilla versions. 
After LogiGAN pre-training, we discard the auxiliary Verifier (discussed in Sec. \ref{sec:discussion}) and employ the Generator only to solve all downstream tasks in a purely end-to-end manner.
For our main experiments, we initialize Generators from both base and large size pre-trained T5 \citep{2020t5}, and Verifier from pre-trained ALBERT-large \citep{DBLP:journals/corr/abs-1909-11942}. We leave discussions of the rest implementation details and hyper-parameter settings of pre-training and downstream fine-tuning in Appendix D.

% \textbf{IV. Replacing Self-Sampling with Retrieved Existing Sentences}
\paragraph{Elastic Search vs. Self Sampling.}
As stated earlier in section \ref{sec:analogy}, candidate pseudo-statements have two possible sources -- they could either be sampled via beam search from the Generator's self-distribution, or could be retrieved from some external resources.
We carry out two variant versions of LogiGAN whose Generator is trained purely from self-sampled sentences as pseudo-statements (\textbf{LogiGAN$_{~base}$(ss)}), and from extra pseudo-statements retrieved from corpus by Elastic Search \cite{Gormley2015ElasticsearchTD} (\textbf{LogiGAN$_{~base}$}(ss+es)). 
For the large model, we use LogiGAN$_{~large}$ (es+ss) as default.
Our database consists of 3.14 million sentences discovered by the corpus construction process, and we keep the top-5 similar retrieved sentences along with self-samples from Generator.  

\section{Experiments} \label{sec:exp}

\subsection{Experimental Results}
% Table generated by Excel2LaTeX from sheet '主实验'
\vspace{-2mm}
\begin{table}[htbp]
  \centering
  \caption{Main results of LogiGAN on 12 downstream tasks (\textit{development sets}).}
  \resizebox{0.9\textwidth}{!}{
    \begin{tabular}{ccccccccc}
    \toprule
    \multicolumn{9}{c}{Multiple Choice \& Classification Datasets} \\
    \midrule
    \multicolumn{2}{c} {Models\,/\,Dataset} & ReClor & LogiQA & RACE  & DREAM & ANLI  & MuTual & \multicolumn{1}{c}{Avg.} \\
    \multicolumn{2}{c}{\small Metrics} & {\small {\small $Acc$}}   & {\small {\small $Acc$}}   & {\small {\small $Acc$}} & {\small {\small $Acc$}} & {\small {\small $Acc$}} & {\small {\small $Acc$}} &  \\
    \midrule
    \multicolumn{2}{c}{Vanilla T$5_{~base}$} & $35.20$ & $27.19$ & $63.89$ & $59.36$ & $44.10$ & $67.38$ & $49.52$ \\
    \multicolumn{2}{c}{LogiGAN$_{~base}$ (ss)} & $40.20$ & $34.72$ & $67.13$ & $63.38$ & $49.50$ & $69.41$ & $54.06$\\
    \multicolumn{2}{c}{LogiGAN$_{~base}$ (ss+es)} & $40.00$ & $37.02$ & $67.27$ & $63.73$ & $49.70$ & $69.98$ & $54.62$\\
    \midrule
    \multicolumn{2}{c}{Vanilla T$5_{~large}$} & $50.40$ & $38.56$ & $78.99$ & $78.98$ & $58.00$ & $76.41$ & $63.56$\\
    \multicolumn{2}{c}{LogiGAN$_{~large}$} & $54.80$ & $40.55$ & $80.67$ & $81.42$ & $63.50$ & $77.88$ & $66.47$\\
    \midrule
    \multicolumn{9}{c}{Generation Datasets} \\
    \midrule
    \multicolumn{2}{c} {Models\,/\,Dataset} & QuoRef & HotpotQA & NarrativeQA & TellMeWhy & SAMSum & XSum & \multicolumn{1}{c}{Avg.} \\
    \multicolumn{2}{c}{\small Metrics} & {\small {\small EM\,/\,F$_1$}} & {\small {\small EM\,/\,F$_1$}}  & {\small Rouge$_L$} & \small Rouge$_L$ & {\small Rouge$_L$} & {\small Rouge$_L$} & \\
    \midrule
    \multicolumn{2}{c}{Vanilla T$5_{~base}$}  & $70.76$\,/\,$74.58$ & $61.11$\,/\,$74.86$ & $48.11$ & $30.03$ & $39.32$ & $29.14$ & $36.65$\\
    \multicolumn{2}{c}{LogiGAN$_{~base}$ (ss)} & $75.02$\,/\,$78.68$ & $62.68$\,/\,$76.14$ & $49.44$ & $31.18$ & $39.92$ & $30.26$ & $37.70$\\
    \multicolumn{2}{c}{{LogiGAN$_{~base}$} (ss+es)} & $74.94$\,/\,$78.40$ & $62.80$\,/\,$76.18$ & $49.46$ & $31.15$ & $40.21$ & $30.27$ & $37.77$ \\
    \midrule
    \multicolumn{2}{c}{Vanilla T$5_{~large}$} & $80.06$\,/\,$83.25$ & $66.11$\,/\,$79.80$ & $51.09$ & $31.42$ & $41.40$ & $31.58$ & $38.87$ \\
    \multicolumn{2}{c}{LogiGAN$_{~large}$} & $81.92$\,/\,$85.25$ & $67.04$\,/\,$80.36$ & $51.79$ & $32.72$ & $43.13$ & $33.49$ & $40.28$ \\
    \bottomrule
    \end{tabular}}
  \label{tab:main}%
  \vspace{-2mm}
\end{table}%
As presented in Table \ref{tab:main}, 
both base and large size PLMs further pre-trained with LogiGAN surpass their vanilla baselines across both discriminative and generative task formats, through a wide scope of downstream tasks requiring general reasoning abilities. 
We can make the following observations:
Among all observed improvements, those on tasks with particular emphasis on logic (ReClor, LogiQA, and ANLI) are most noticeable.
These positive results manifest the effectiveness of LogiGAN in injecting logic ability into PLMs, while testifying to our primary hypothesis that logic ability is fundamental to general reasoning as well.
This conclusion answers the two questions in the intro section \footnote{Is logic ability obtained for free from MLM? Could it be further learned
beyond language acquisition?}, suggesting that randomized MLM pre-training might fall short in endowing language models with logic ability, and a logic-targeted pre-training approach like LogiGAN may further assist logic learning beyond language acquisition.
Furthermore, extra retrieved pseudo-statements (ss+es) bring some additional performance improvement compared with the pure self-sampling (ss) LogiGAN variant, revealing the important role of pseudo-statements' \textit{diversity} in adversarial training.

\subsection{Ablation Study and Analysis}

% Table generated by Excel2LaTeX from sheet 'Ablation'

Observing the apparent performance enhancement, we now aim at pinpointing the truly functional components of LogiGAN through ablation studies and deriving the origins of observed improvements.
For fair comparison purposes, we hold all pre-training and downstream settings (including hyper-parameters, implementation designs, and evaluations) unchanged from full LogiGAN.
All variations are initialized from T$5_{base}$, and we report performance variance on $7$ datasets. 

\definecolor{Gray}{gray}{0.9}
\begin{table}[htbp]
  \centering
  \caption{Ablation Results on $7$ datasets. The last column shows average performance variance, along with relative percentage improvement against vanilla T$5_{base}$ as the baseline.}
  \resizebox{0.9\textwidth}{!}{
    \begin{tabular}{cccccccccc}
    \toprule
    \multicolumn{2}{c}{Models\,/\,Dataset} & ReClor & LogiQA & RACE  & DREAM & ANLI & QuoRef & NarrativeQA &{---} \\
    \multicolumn{2}{c}{\small Metrics} & {\small $Acc$}   & {\small $Acc$}   & {\small $Acc$}   & {\small $Acc$} & {\small $Acc$} & {\small EM\,/\,F$_1$} & {\small Rouge$_L$} & {\small Average} \\
    \midrule
    \multicolumn{2}{c}{Vanilla T5$_{~base}$} & $35.20$  & $27.19$ & $63.89$ & $59.36$ & $44.10$ & $70.76$\,/\,$74.58$ & $48.11$ & $49.80_{(+0.0\%)}$ \\
    \rowcolor{Gray}
    \multicolumn{2}{c}{LogiGAN$_{~base}$ (ss+es)} & $40.00$  & $37.02$ & $67.27$ & $63.73$ & $49.70$ & $74.94$\,/\,$78.40$ & $49.46$ & $54.59_{(+9.6\%)}$ \\
    \midrule
    \multicolumn{2}{c}{I. Random Sentence} & $36.00$ & $30.56$ & $61.26$ & $58.15$ &  $45.40$ & $70.96$\,/\,$74.50$ & $48.38$  & $50.10_{(+0.6\%)}$ \\
    \multicolumn{2}{c}{II. MLE Logic Pre-train } & $38.80$  & $35.02$ & $64.55$ & $61.71$ & $46.00$ & $73.61$\,/\,$76.96$ & $49.30$ & $52.71_{(+5.9\%)}$ \\
    \multicolumn{2}{c}{III. Iterative Multi-task} & $37.20$  & $34.25$ & $64.01$ & $62.06$ & $46.20$ & $71.67$\,/\,$75.14$ & $49.15$ & $52.08_{(+4.6\%)}$  \\
    \bottomrule
    \end{tabular}}
  \label{tab:ablation}%
  \vspace{-5mm}
\end{table}%

\textbf{I. Random Masked Sentence Prediction Pre-training.}~
To explain the observed improvements, our first hypothesis is: Models harvest \textit{extra linguistic ability} from masked \textit{statement} prediction compared with masked \textit{token (or span)} prediction.
Quite intuitively, filling entire sentences with complete subject-predicate structures might put additional demands on models to capture more abundant syntactic information beyond the coverage of masked token (or span) prediction.
Since LogiGAN involves recovering masked \textbf{\textit{sentences}}, it is then necessary to determine to what degree, if any, that the observed performance gain is attributable to models' plausible linguistic ability improvement. 
We therefore carry out a variant pre-training where the prediction objects are \textit{randomly masked sentence}.

% Since logic reasoning are primarily dependent on linguistic abilities -- language as lower infrastructures, and logic reasoning as upper establishments -- our important first step will be to disentangle linguistic ability from logic ability.
 
Results (shown in Table \ref{tab:ablation}) displays that masked sentence prediction training barely brings improvement against the vanilla baseline.
This suggests it is unlikely that masked sentence prediction empowers PLM trained from masked token prediction significantly better linguistic ability, nor likely that the extra pre-training corpus per se significantly raises the performance.
Therefore, we reject the first hypothesis and conclude that observed improvements should derive from somewhere else.
% Accounting overlaps between BookCorpus and T5'native $750GB$ C4 corpus, as well as the low marginal benefits of $2GB$ extra pre-training corpus per se --
% the improvement should therefore derive elsewhere.

\textbf{II. MLE-only Logic Pre-training.}~
Our second hypothesis is that logic-guided masked statement prediction enhances models' intrinsic ability of logical reasoning, thereby lifting the downstream performance.
Having addressed the potential impact of learning randomized complete sentence generation, we next aim to check how learning logic-targeted statement generation affects models' behavior.
We ablate the entire adversarial training process, and train models to perform maximum likelihood estimation (MLE) with teacher-forcing loss only on masked-out logical statements.

% Since logic indicators provides reliable marks for such logic phenomena, we argue training models to predict the statements after logic indicators has dual function by its nature:
% (i) a pure linguistic aspect that is also captured by random masked sentence prediction;
% (ii) an advanced reasoning logic aspect that is deeply embedded in the semantics, instead of the superficial grammar or syntax of these statements.
% Going beyond \textit{I - Random Masked Sentence Prediction Pre-training} that is designed to explore impact of (i), we here investigate the impact of (ii) through carrying out pure generative logic pre-training.

Results \ref{tab:ablation} of MLE-only logic pre-training reveals quite a notable improvement across almost all datasets against both vanilla baseline and I., suggesting that learning to generate logical statements indeed injects extra abilities into the model.
Since results of I. eliminate the possibility that models harvest stronger linguistic abilities from complete sentence prediction, it is safe to partially ascribe the better downstream performance to models' enhanced ability in modeling logical reasoning. 
This reveals the relative orthogonality between logic ability and models' inherent linguistic ability, suggesting that logic ability could be enhanced through further logic-targeted pre-training.
 
\textbf{III. Iterative Multi-task Pre-training.}~
Since II. only partially explains the observed improvements, here is our last hypothesis: the adversarial training procedure of LogiGAN explains the unexplained rest part beyond the coverage of II.  
% This hypothesis seems plausible because critical discrimination logical coherency between premises and conclusions and conditional conceiving of logical from scratch appears to be distinctive from each other.
Here a multi-task pre-training with both generation and verification tasks will be the most natural intermediate setting between the \textit{single-model generation-only setting} of II. and LogiGAN's \textit{dual-model adversarial setting}. 
However, since the verification task relies on Generator's self-sampled statements, we adopt an iterative self-critic pre-training manner following \cite{nijkamp-etal-2021-script}.
Unlike typical multi-tasking training that simultaneously carries different tasks and then sums the losses,
our generation and verification tasks happen alternately
\footnote{Verification is formulated as a generation task -- model outputs natural language token ``good'' and ``bad''.}.
% After the initial warm-up just as done in GAN, the Generator produces exactly ONE answer for each training example from Verifier training corpus.
% The rest of the training is similar to the training process of GAN described in \ref{gan-train}.

Surprisingly, the iterative multi-task pre-training barely brings any positive effects to models' downstream performance compared with II.
One possible explanation for this might be that
the drastically different mechanisms between the verification and generations task intervene with each other,
making the single-model \& multi-task setting non-beneficial. 
Now that we have confirmed that an extra verification task fails to explain the rest improvement,
we can accept our final hypothesis and conclude that it is indeed the adversarial mechanism between the Generator and Verifier that truly facilitate learning of logical reasoning, thereby further improving the downstream performance beyond II.

% \section{Discussion} \label{sec:discussion}
% % In the discussion section, we share insights that might inspire future research.
% \paragraph{Adversarial Training Might Assist Generative Downstream Tasks}
% In our main experiments, we discard the Verifier and solve downstream tasks with the Generator only.
% According to our other empirical results, directly applying Verifier in downstream tasks leads to performance drop, which might be attributable to the strong inductive bias from the binary-classification pre-training task.
% However, the pre-trained Verifier could still be useful:
% some previous works \citep{DBLP:journals/corr/abs-2109-03034, cobbe2021gsm8k} reveal that the Verifier can be used for ranking multiple generation results, thereby effectively enhancing overall downstream accuracy.
% % They also suggest that a generator-verifier architecture can be more parameter-efficient than a single generator. 
% However, in their paradigm, the information propagates unidirectionally from the Generator to the Verifier, and the Generator cannot directly benefit from the Verifier's discriminative feedback.
% In contrast, our LogiGAN adversarial training paradigm surmounts the non-differentiable obstacle and could potentially enlighten a new paradigm of both pre-training and downstream fine-tuning.

\section{Discussion} \label{sec:discussion}
% In the discussion section, we share insights that might inspire future research.

\paragraph{A Psycholinguistic Interpretation of Logic-oriented MLM}
From the first glance, the idea of Logic-oriented MLM seems to be naive and simply. However, we argue that to fully appreciate the value and potential of logic-oriented MLM, going beyond the superficial appearance of masked text and touching down to their underlying psycholinguistic essence is necessary.

It is neither the linguistic patterns of the masked-out text, nor the masking technique to corrupt them that makes logic-oriented MLM (and its potential follow-ups) unique. What truly matters is the distinctive \textbf{\textit{cognitive processes}} proceeding in the minds of writers when they put down different pieces of text -- language is a window into human minds \citep{pinker2007stuff}.

Consider the following examples when a human is filling in the [MASK]’s:
\begin{itemize}
    \item[(1)] “19 + 69 = [MASK]." (Numerical cognition).
    \item[(2)] “Windows is founded by [MASK].” (Declarative memory retrieval).
    \item[(3)] “Socrates is a mortal, so he will eventually [MASK]” (Logical reasoning).
    \item[(4)] “If I feed my dog more than 2 treats per day, it will get [MASK].” (Causal inference).
    \item[(5)] "A crow immediately stands out in swans because it's [MASK]."  (Common sense reasoning).
    \item[(6)] “Mike deeply bows to his teacher to show his [MASK].” (Social perception).
\end{itemize}

Though answers to these [MASK]’s are similar in string length, they nevertheless involve  different information pathways and substantially distinctive cognitive processes in writers' minds.

Logic-oriented MLM shines in that it consistently captures exactly one type of such cognitive process, and trains LMs to model humans’ logic reasoning mechanism, which is well beyond modeling language per se. On the contrary, Randomized MLM does not capture consistent underlying cognitive processes, which could significantly lower LMs' efficiency of learning advanced intelligence mechanisms beyond language itself. The empirical effectiveness of Logic-oriented MLM provides positive evidence for the practicability of this paradigm, suggesting that LMs might be able to learn various advanced human cognitive processes other than logical reasoning via a similar approach. Combined with LogiGAN's natural analogy of humans' learning-thinking mechanism during logic development, LogiGAN made an encouraging attempt to unify cognitive modeling and language model pre-training.

\paragraph{Adversarial Training Might Assist Downstream Generation Tasks.}
Although in our experiments, we discard the Verifier and solve downstream tasks with the Generator only, some previous works \citep{DBLP:journals/corr/abs-2109-03034, cobbe2021gsm8k} reveal that the Verifier can be used for ranking multiple generation results, thereby effectively enhancing overall downstream accuracy.
% They also suggest that a generator-verifier architecture can be more parameter-efficient than a single generator. 
However, in their paradigm, the information propagates unidirectionally from the Generator to the Verifier, and the Generator cannot directly benefit from the Verifier's discriminative feedback.
In contrast, our LogiGAN adversarial training paradigm surmounts the non-differentiable obstacle and could potentially enlighten a new paradigm of both pre-training and downstream fine-tuning.

\paragraph{Improving Logical Pre-training.}
% Our paper demonstrates that the logic ability can be further enhanced with continual pre-training, and adversarial training facilitate learning of logical reasoning.
Our paper demonstrates that PLMs' logic ability can be further enhanced beyond their inherent linguistic ability, and adversarial training may bring extra benefits beyond the learning of logic-targeted masked statement prediction.
However, our heuristic-based approach to identifying logical phenomena in a text corpus and the single mask prediction setting can be further improved.
Logic recognition methods with higher recall and better unsupervised task designs (e.g., \textit{logical indicator prediction}, or \textit{logic-guided sentence shuffling}) are worthwhile to explore in the further work.
Besides, since we are adopting a general domain pre-training corpus (i.e., \textit{BookCorpus}) with bare emphasis on logic, understanding the impacts of extending pre-training to the domain-specific corpus (e.g., law corpus) or others emphasizing logical reasoning is also substantial.

% \subsection{Our Primary Assumption of Informal Logic Pre-training}
% Our primary assumption for conducting the informal logic pre-training is:
% the valid conclusion draw-able following a given set of premises is a certain distribution of semantics, and there does not necessarily exist a single ``ground truth conclusion'' as is case of formal logic.
% For example, in the case ``Bob recently got obsessed with junk food. Today he went on the scale and was shock to observe his huge weight gain. He makes up his mind to lose weight. Therefore, \texttt{[MASK]}''.
% Conclusion-1 ``he decides to go on a diet'' and conclusion-2 ``he decides to exercise from today on''. might both be reasonable.
% Therefore, in contrast with the the formal logic scenario where a single gold answer is necessarily deducible based on given premises, our pre-training aims to maximize models' probability of producing conclusions from the logically reasonable semantic spaces.
% ndom sentence pre-training,

\section{Related Works}

\paragraph{Generative Adversarial Training in NLP.}

Unlike conventional GAN \citep{https://doi.org/10.48550/arxiv.1406.2661, DBLP:journals/corr/MirzaO14, DBLP:journals/corr/ZhuPIE17} that generates continuous output such as images, sequential GAN generates discrete sequences via non-differential searches.
This makes feedback from the discriminator not propagatable to the generator.
To tackle this challenge,
\textbf{SeqGAN} \citep{10.5555/3298483.3298649} borrows an idea from reinforcement learning, treating each output token as a single action, and estimates token-wise policy gradient via Monte Carlo search. \textbf{RankGAN} \citep{Lin2017AdversarialRF} adopts a similar approach but breaks the binary-classification assumption of discriminator task design, and a ranker provides feedback to the generator.
Their generator attempts to generate verisimilar sentences to deceive the ranker into ranking synthetic sentences higher over multiple human-written ones.
In our scenario, however, the gold ranking is hard to determine because measuring which statements are more logically consistent w.r.t. context than others is non-trivial, and multi-gold cases are possible.
% each context is paired with exactly one human-written statement, and it is unclear how the rest statements acquired by generation or retrieval should be ranked, since precisely measuring the logical consistency of a context-statement pair is non-trivial.
While successfully enabling communication between generator and discriminator, the original designs of SeqGAN, RankGAN, as well as other works such as \citep{DBLP:journals/corr/abs-1709-08624, https://doi.org/10.48550/arxiv.1801.07736, DBLP:journals/corr/abs-1811-02549, Rekabdar2019GenerativeAN}, generally formulate text generation as a sequential action decision problem, thereby involving heavy sampling for policy gradient estimation, and are sensitive to the length of the target sequence.
Since large-scale pre-training (with arbitrary target length) puts a high demand on scalability and computational efficiency, the above approaches are not readily applicable in our scenario.
Furthermore, previous work leverages adversarial training to \textit{improve qualities of generated examples}, whereas our focus is on \textit{enhancing models' intrinsic logic ability}.

% A ranker is trained to rank based on verisimilitude. The goal of their generator is to generate verisimilar sentence such that the ranker is fooled to assign higher rank to the fake sentence over human-written sentences. However, since in our unsupervised pre-training scenario, we have only one real answer, with multiple fake examples, it unclear how the ranking of should be ranked.

A recent work, \textbf{AR2} \citep{DBLP:journals/corr/abs-2110-03611}, leverages adversarial training to improve dense document retrieval. 
With a retriever-ranker architecture, the learning objective of retriever is to maximize the agreeableness between its own score assignment and that of the ranker for input documents.
This is conceptually similar to LogiGAN, as our Generator also aims at reaching consensus with Verifier.
However, AR2 does not fall into the sequential GAN paradigm, since it does not involve any sequential text generation, and there is no non-differentiable barrier between the retriever and ranker.
% like beam-search that prevents the direct gradient propagation 

\paragraph{Pre-training for Reasoning Ability Improvement.}
Previous works have extensively investigated the possibility of injecting specific type of reasoning via pre-training, such as
numerical \citep{ DBLP:journals/corr/abs-2004-04487,DBLP:journals/corr/abs-2107-07261, DBLP:journals/corr/abs-2201-11473}, commonsense  \citep{zhong2019improving,DBLP:journals/corr/abs-2004-14074, DBLP:journals/corr/abs-2101-04966},
formal logic \citep{wang2021logic, DBLP:journals/corr/abs-2201-11473}, 
multi-hop \citep{deng2021reasonbert,zhong2022reasoning}, 
and tabular \citep{DBLP:journals/corr/abs-2107-07653} reasoning.
Different from them, LogiGAN focuses on logic reasoning, which plays a fundamental role in general reasoning via natural language.

\section{Conclusion}
% Upon analysis of logic and reasoning, we hypothesize that improving the logic ability of PLMs might benefit various downstream tasks requiring general reasoning.
% Correspondingly, we propose LogiGAN, an unsupervised adversarial pre-training framework for logic ability enhancement.
% LogiGAN circumvents the non-differentiable challenge of sequential GAN with a novel adversarial mechanism based on the Generator-Verifier scoring consensus.
% This design fits well with the multi-gold property of logical reasoning and enables large-scale pre-training with arbitrary target length.
% The pervasive improvement of both base and large-sized PLMs on various downstream tasks requiring general reasoning abilities confirms the importance of logic in reasoning processes.
% Ablation studies on LogiGAN components further reveal
% the relative orthogonality between linguistic and logic abilities, as well as the effectiveness of adversarial pre-training paradigm for logic enhancement.

In this work, we hypothesize that (i) logic ability plays a key role in a wide scope of tasks requiring general reasoning; and (ii) PLMs' logic ability can be further improved beyond their original linguistic ability.
We correspondingly propose LogiGAN, an unsupervised adversarial pre-training framework for logical reasoning enhancement.
LogiGAN circumvents the non-differentiable challenge of sequential GAN via a
novel Generator-Verifier scoring consensus mechanism, and enables large-scale pre-training with longer target length.
Extensive experiments and ablation studies reveal the effectiveness and functional components of LogiGAN, providing evidence to our major hypothesis.

% Ablation studies on LogiGAN components further reveal
% the relative orthogonality between linguistic and logic abilities, as well as the effectiveness of adversarial pre-training paradigm for logic enhancement.

% Upon analysis of logic and reasoning, we hypothesize that improving the logic ability of PLMs might benefit various downstream tasks requiring general reasoning.

% LogiGAN circumvents the non-differentiable challenge of sequential GAN with a novel adversarial mechanism based on the Generator-Verifier scoring consensus.
% This design fits well with the multi-gold property of logical reasoning and enables large-scale pre-training with arbitrary target length.
% The pervasive improvement of both base and large-sized PLMs on various downstream tasks requiring general reasoning abilities confirms the importance of logic in reasoning processes.
% Ablation studies on LogiGAN components further reveal
% the relative orthogonality between linguistic and logic abilities, as well as the effectiveness of adversarial pre-training paradigm for logic enhancement.

\bibliography{anthology,custom}
\bibliographystyle{nips}
\appendix

\section{Thinking Straight about Reasoning in NLP} \label{app:reasoning}

The following argument is tentative and is deduced with bounded rationality,
and is \textbf{for communication purposes only}.
We realize and well acknowledge that researchers observing different sets of evidence could hold fundamentally different but reasonable views from ours.

\subsection{Conundrums of Reasoning}
Along with the increasing interest in reasoning, multiple reasoning terms are proposed --  hybrid reasoning, commonsense reasoning, numerical reasoning, multi-hop reasoning, and unspecified general reasoning, to name a few.
However, among all these scattered and distinctive types of reasoning, what is varying?
What essence remains constant regardless of the variety of forms?
If we were to arrange these reasoning into a hierarchical structure much like a biological taxonomy or to group them into categories, what standard should we follow?
How should we put boundaries between each type of reasoning?
To the best of our knowledge, few works from the NLP community articulated these queries well.
There seems to be a \textbf{conceptual conundrum} of reasoning.

Moreover, the ethereal and shapeless nature of reasoning makes it not as visible or concrete as tokens or spans that are readily accessible to the masked language modeling pre-training paradigm.
How could we then systematically inject reasoning ability into models via pre-training?
What is the nature of this ethereal ability are we truly pursuing?
More fundamentally, is pre-training the correct way to add reasoning abilities into language models?
Should reasoning abilities be acquired during the pre-training stage, or should it be subsequently tackled by outsourcing symbolic modules (i.e., with neural-symbolic models)?
There seems to be a \textbf{methodological conundrum} of reasoning.

These questions are non-trivial to answer, and these morals are hard to tell.
However, it is quite unlikely that we can solve the reasoning challenge before articulating what the challenge truly is.
% these challenges of reasoning are what we willing to accept, to further investigate the possibilities of stronger intelligence.
% With deeper inquiries guided by the curiosity, we are able to reach a plausible answer: \textbf{logic reasoning}, the systematic approach of deriving statements faithful to the truth, lies at the most fundamental level, and is shared across all sorts of reasoning.

\subsection{Reasoning in our Sense} \label{app:reasoning-def}
As defined in the introduction section, reasoning (via natural language) is an \textit{inferential process where an unstated statement is drawn based on several presented statements.}
Specifically for deductive and inductive reasoning, a conclusion is drawn from provided premises under the guidance of logic.
Here is an example of the most trivial cases of such reasoning: given premises ``Bob's daughter is called Lily. Lily is now $3$ years old.'', concluding ``Thus, Bob's daughter is $3$ years old.''  requires shallow synthesis between exactly-matched subject and predicate terms.

However, in complex machine reading comprehension tasks, such single-step synthesis can be arbitrarily long-chained (e.g., requires $5$ syntheses, and 
combined with semantic-invariant linguistic transformations such as synonym replacements and syntactic transformations.
As a result, reasoning over long and rhetorically sophisticated articles becomes non-trivial, demanding highly on both linguistic and logic abilities.

Apart from linguistic transformations, synthesizing among statements usually requires specific synthesis rules of statements.
For example, the synthesis rule of degree comparison ``Elephants are larger than dogs. Tigers are larger than dogs.'' $\rightarrow$ ``Elephants are larger than dogs.'' does not apply to the case
``In Pokemon, the water type is strong against the fire type, and the fire type is strong against the grass type.''
since the conclusion is guided by the same rule ``Therefore, the water type is strong against the grass type.'' fallaciously contradicts reality.
Putting into broader domains, mathematics (e.g., arithmetic operators, set operators), formal logic (e.g., quantifiers, logic operators), and most artificial symbolic systems implement their own synthesis rules as standards of correctness.
Therefore, while the forms of reasoning remain relatively invariant, the underlying synthesis mechanism could be drastically different from system to system.
Among all these symbolic reasoning, the reasoning in NLP focuses primarily on reasoning via natural language -- i.e., linguistics as synthesis rules.

\subsection{General Reasoning and Specialized Reasoning}
Following the definition of reasoning we make in the previous subsection, we are now able to tentatively categorize all investigations of reasoning in NLP into two families:
\textbf{Specialized Reasoning and General Reasoning.} We discuss them separately below:
\paragraph{I. Specialized Reasoning} can be further divided into sub-categories:

\textit{\textbf{(a) Reasoning requiring special way of premise extraction}}, such as multi-hop reasoning \citep{yang2018hotpotqa}, tabular reasoning \citep{zhong2020logicalfactchecker}, and hybrid reasoning \citep{chen-etal-2020-hybridqa}.  
The foremost assumption in this scenario is that the input context has already provided all premises necessary to draw targeted conclusions.
If we humans are trying to answer a question correctly (i.e., with both correct answers and reasons),
we will have to first seek back-and-forth across multiple documents or paragraphs (multi-hop) or rows and columns (tabular) to extract premises for answering this question.
Based on these extracted premises, we then follow the logic rules and synthesize a conclusion from these premises to answer the question.
Not drastically different from humans, in such complex reasoning scenarios where spurious patterns are mostly unreliable, machines will also have to identify the necessary premises correctly to reach correct conclusions.
With all relevant and necessary premises extracted, the rest of the reasoning are reducible to general reasoning described below in II.
% Suppose necessary premises are already gathered and redundant information are filtered out, the rest 
% Due to the complex nature of these reasoning scenarios, reliance on spurious pattern will likely to results in wrong answers.
% Here we draw a analogy between human attention mechanism and that of transformers.
% Fellow research from the cognitive psychology has long confirmed that eye gaze cues \citep{frischen2007gaze} and eye movements \citep{henderson1992visual} provides reliable explanation to the variance of human attention.
 
\textbf{\textit{(b) Reasoning requiring external premises}}, such as numerical reasoning \citep{Dua2019DROPAR}, symbolic reasoning \citep{zhong2021ar}, domain-specific reasoning \citep{wang2022lsat,gao2021open}, and commonsense reasoning \citep{sap-etal-2020-commonsense}.
This family of reasoning requires external knowledge that is harder to be acquired via typical language modeling pre-training.
Here we emphasize that knowledge can either be \textbf{\textit{declarative}} or \textbf{\textit{procedural}}, following the theory from psychology \citep{ten1999procedural, ullman2001neurocognitive}.
For example, the commonsense knowledge ``There are 365 days in a year on earth.'', or knowledge of historical events can be primarily declarative (i.e., can be articulated with language).
Other not readily articulable knowledge, such as swimming, and performing complex arithmetics by hand (e.g. $1969 * 331$) are primarily procedural.

In fact, humans need extra learning to acquire specialized knowledge. For example, we do not acquire knowledge of mathematics or domain expertise for free from language learning.
Just as ``import PyTorch'' is a necessary dependency to implement our neural models in Python, we have to implicitly invoke established mathematics rules to perform calculations,
and invoke domain-specific knowledge or laws as premises to make scientific arguments.
Machines are generally no different.
One special type of premise is commonsense, a large set of knowledge that human effortlessly obtains from multi-modal life experiences.
% PLMs, in contrast, much resemble brains in vats, are not fortunate enough to have the commonsense premises merely from MLM pre-training.
Suppose all necessary premises are readily invoked and stated in natural language, the rest of the reasoning proceeds just as general reasoning, which is discussed below in II.

\paragraph{II. General Reasoning}

General reasoning covers a broad unspecified form of reasoning that involves recognizing and understanding relevant concepts framed in plain text as premises and then synthesizing them into conclusions.
Typically, it assumes stated premises are mostly self-contained (i.e., sufficient for drawing certain conclusions), and require little or no external knowledge to be additionally invoked beyond the given context.
Moreover, since premises are presented in a plain text format, general reasoning does not require special ways of extracting premises from context (e.g., structured data such as tables).
The general reasoning is pervasive in tasks emphasizing natural language understanding (NLU), such as machine reading comprehension \citep{gao2020discern} and natural language inference \citep{DBLP:journals/corr/abs-1910-14599}.
General reasoning also serves as an underlying foundation of other specialized reasoning tasks, since most of them can be conditionally reducible to general reasoning, as discussed earlier.
 
During general reasoning processes, premises can be organized into a logic chain via procedures such as alignment of semantically similar concepts, understanding relations among sentences, and synthesizing sub-conclusions from specific subsets of premises.
\textbf{Logic} is the systemic set of principles created for providing
correctness assurance to conclusions inferred following such logic chains,
upon examination of coherence and consistency of these chains \footnote{Notice that premises from given context are assumed to be true when solving NLP downstream tasks.}.
Invocation of logic is therefore necessary for reasoning to be correct (i.e., drawing correct conclusions from correct reasons) -- although humans and machines usually carry this out implicitly.
Conclusively, logic is the core engine for general reasoning.
\textit{\textbf{This completes our analysis on logic and reasoning from the introduction section.}}

% Suppose \textbf{\textit{all}} necessary premises are stated,
% it is then technically possible to formalize these stated premises into a nose-to-tail compact form, with the predicate term of ${(i-1)}^{th}$ statement exactly matches the subject term of  $i^{th}$ statement.
% This formalization is carried via semantic-invariant linguistic transformations, such as sentence breaking, reordering, synonym replacement, corefernece resolution, syntax tree rotation, and modality shift.
% The resulting chain of premises is reduced to the trivial case of reasoning mentioned earlier in \ref{app:reasoning-def}.
% Albeit being technically possible, such formalization process is almost always unnecessary for human or PLMs to perform general reasoning in real practices.
% Thanks to linguistic abilities, the above semantic-invariant linguistic transformations could happen implicitly, after which a final conclusion will be synthesized.

\paragraph{Some Comments on our Categorization}
Regarding \textbf{I-(b)}, we realize it might be less intuitive or even controversial to consider procedural knowledge as ``premises''.
In contrast with crystallized declarative knowledge that resembles more of \textit{\textbf{String type variables}} in computer programs,
fluid procedural knowledge resembles more of \textit{\textbf{functions}} -- although functions have well-defined function bodies while procedural knowledge is usually beyond words.
Therefore, tasks requiring external declarative knowledge (e.g., commonsense, domain-specific expertise) and procedural knowledge (e.g. symbolic calculations formal logic, arithmetics) might be further divided into two sub-categories.

Apart from the above, our coverage bounds to deductive and inductive reasoning via natural language.
Abductive reasoning is the reverse reasoning process where premises are hypothesized to explain or support some stated conclusion,
and symbolic reasoning is beyond the coverage of linguistics.

\section{Bridging the Gap between Pseudo and Logical Inconsistency} \label{app:gap}
As mentioned in the footnote of  Sec. 3 (The Adversarial Training Framework) from the main text, 
there is a gap between pseudo-statements that are either self-sampled from the Generator or retrieved and logically inconsistent statements.
i.e., there might be logically consistent statements that should have received a positive label when training the Verifier assigned with a negative label since they are pseudo.
This could potentially introduce noise signals in the training of the Verifier.

To bridge this gap, we propose a trick that leverages off-the-shelf Natural Language Inference (NLI) models to make judgment on the textual entailment between the pseudo-statement and the ground-truth statement.
Our basic intuition here is that: suppose a pseudo-statement implies or is implied by the ground-truth statement, then this pseudo-statement is also expected to be logically consistent w.r.t. the original context, just as the ground-truth one.

For example, in the example of Fig. 1 (LogiGAN Overview), with context, ``\textit{All men are mortal, and Socrates is a man. Therefore}, \texttt{[MASK]}.'', the ground-truth statement ``\textit{Socrates is mortal.}'' implies the self-sampled pseudo-statement ``\textit{Socrates will eventually die}.'' from Generator, vice versa.
Hence this pseudo-statement should receive a positive label (i.e., $y=1$) for training Verifier, who learns to discriminate logical consistency of statements w.r.t. input context, instead of merely capturing fake examples.
In contrast, the first retrieved pseudo-statement ``\textit{a mortal can never be a god.}'' does not entail, nor is entailed by, the ground-truth statement. 
Therefore this pseudo-statement will have a negative label for training Verifier.

Technically, we employ ``ynie/albert-xxlarge-v2-snli\_mnli\_fever\_anli\_R1\_R2\_R3-nli'' from Huggingface, a well-trained NLI model (denoted as $\mathcal{F}$), to determine the textual entailment relationship between ground-truth and pseudo-statements.
The process can be formally expressed as:
$$
e(s^+, s^-) = \max (~\mathcal{F}(s^+, s^-)~,~ \mathcal{F}(s^-, s^+)~),
$$
where $e(s^+, s^-)$ represents the entailment score between the ground-truth statement $s^+$ and a pseudo-statement $s^-$.
To determine the final label for training Verifier, we set a hard threshold of $0.50$ -- above results in $y=1$ below turns to $y=0$.
According to our statistical study, this extra NLI mechanism flips around $12\%$ of the pseudo-statements (whose default labels are negative) to $y=1$.
With our human evaluation, the flipped pseudo-statements are indeed logically consistent w.r.t. the original context in most cases.

As an emphasis, the NLI model and the Verifier are making judgment on different things.
The NLI does not consider the original context and merely judges the entailment of a statement pair, whereas the Verifier judges one statement's logical consistency w.r.t. the context at a time.
The signal from the NLI serves noisy and indirect supervision in Verifier's learning process, and we are leveraging the intrinsic denoising ability of large pre-trained language models.
Ablation of this mechanism results in a minor performance drop, but not significant enough to include in our main ablation study, so we are omitting this in the main text.

\section{List of Logic Indicators} \label{app:indicators}
\paragraph{Conclusion Indicators (41 in total):}
therefore, thereby, wherefore, accordingly, we may conclude, entails that, hence, thus, consequently, we may infer, it must be that, whence, so that, so, it follows that, implies that, as a result, it can be inferred that, suggests that, can conclude, proves that, it can be shown, as a conclusion, conclusively, which implies that, for that reason, as a consequence, on that account, that being said, in conclusion, to that end, for this reason, on account of, because of this, that being so, because of that, ergo, in this way, in this manner, in such a manner, by such means.

\paragraph{Premise Indicators (20 in total):}
 since, on account of, considering, because of, because, due to, now that, in order, as indicated by, because, may be inferred from, given that, owing to, by virtue of, owing to, on account of, in view of, for the sake of, thanks to, reason that.
 
 \section{Implementation Details} \label{app:details}
\textbf{LogiGAN Details.}~~
We randomly sample 2 million and 0.5 million for source training examples of Generator and Verifier, respectively (i.e., $M = 2$ million, $N = 0.5$ million in Algorithm 1 in the main text). 
We partition the source Generator training corpus into to two $1$ million subsets for warmup and GAN training (i.e., $M_\alpha = M_\beta = $ $1$ million). 
In each iteration of GAN training, $10\%$ or $0.05$ million and $0.1$ million examples are sampled for Verifier and Generator training (i.e., $m$ = $0.05$ million, $n$ = $0.1$ million). 
The warm-up epoch $E$ is set to $5$, whereas the max number of GAN iteration is set to $10$.
For our main experiments, we initialize the Generator from pre-trained ``google/t5-v1\_1-base'' or ``google/t5-v1\_1-large'', and the Verifier as ``albert-large-v2'' from HuggingFace \citep{wolf-etal-2020-transformers}. 

We use 8 V100 GPUs for model training, and set the maximum iterations of adversarial training as 15, and set batch size as 8 and 64 for Generator and Verfier training.
During Generator training, we put both the only one ground-truth statement, and 5 candidate statements for each instance within the same batch.
By default, we adopt learning rate as 5e-5, 1e-5 for the training of Generator and Verifier during adversarial process, respectively.

 \textbf{Downstream Details.}~~ 
 Our tested downstream datasets mainly belong to two types: generation-based datasets (like extractive QA, abstractive QA, summarization, etc.), and classification datasets (like natural language inference, and multiple-choice QA). We introduce how we process the inputs and outputs for each downstream tasks and the hyper-parameters for fine-tuning.

 For the generation-based datasets, we adopt simple hard prompts to write the task input. For example, for the generative QA task, we formulate the input with ``\textit{The question is: \{question\}. The context is: \{context\}, please give the answer}''. 
 For the classification datasets, with the context (optionally question and options) given as the inputs, the target sequence is one of the options, like ``{entailment}'' for NLI datasets or one candidate answer for MRC datasets. For example, for the multiple-choice QA task, we prompt the input as ``\textit{The question is: \{question\}}. The options are: \{options\}. The context is: \{context\}, please select the best option.'', where the output is the specific content of the option.
 We make the final choice of option by selecting the option with the highest text similarity score calculated by the model output and the context of each option.
 
During fine-tuning, we don't perform exhaustive parameter searches for each task. We adopt the learning rate as either 1e-4 or 5e-5 for each task, depending on which one will lead to stable performance. We adopt 8 V100 GPUs for fine-tuning and set batch size as 32 for the base model, and 8 for the large model.

\section{Few-shot Experiments}

% Table generated by Excel2LaTeX from sheet 'Few-shot'
\begin{table}[htbp]
  \centering
  \caption{LogiGAN Few-shot Setting Performance.}
  \label{tab:few-shot}
  \resizebox{1.0\textwidth}{!}{
    \begin{tabular}{ccccccccccc}
    \toprule
    \multicolumn{2}{c}{Models/Datasets} & RACE  & DREAM & ReCLor & LogiQA & ANLI  & NarrativeQA & $\alpha$NLG & xsum & samsum \\
    \multicolumn{2}{c}{\small metrics} & {\small $Acc$}  & {\small $Acc$} & {\small $Acc$}  & {\small $Acc$} & {\small $Acc$} & {\small Rouge$_L$} & {\small Rouge$_L$} & {\small Rouge$_L$} & {\small Rouge$_L$} \\
    \midrule
    \multicolumn{2}{c}{Vanilla T$5_{~base}$} & $25.27$ & $34.09$ & $26.20$ & $23.34$ & $33.50$ & $5.86$ & $10.90$ & $13.67$ & $13.52$ \\
    \multicolumn{2}{c}{Random Sentence} & $25.26$ & $32.50$  & $26.80$ & $25.03$ & $33.70$ & $16.79$ & $18.28$ & $15.39$ & $26.52$ \\
    \multicolumn{2}{c}{LogiGAN-es$_{~base}$} & $26.28$ & $35.05$ & $28.40$  & $26.57$ & $35.60$ & $18.83$ & $20.80$  & $17.34$ & $27.93$ \\
    \bottomrule
    \end{tabular}}
    
\end{table}%
 
As a supplementary experiment, we also explore LogiGAN's impact under a data-scarce setting.
For each dataset, we randomly select 32 samples and fine-tune models until convergence.
To decouple the complete sentence generation and logic learning in the logic-targeted masked statement prediction process, we also add the random sentence pre-training setting into the comparison.
This could reveal to what extent the performance variance is explained by the linguistic logic ability improvement separately.
Results of \label{tab:few-shot} provide more evidence for our hypothesis that logic ability and linguistic ability have non-overlapping components.

\section{LogiGAN Training with Policy Gradient}
This is to show the overall optimization goal of the adversarial training (Eq. 2 in the main):
$$
J^{\mathcal{G}^{*} \cdot \mathcal{V}^{*}}=\min _{\theta} \max _{\phi} \mathbb{E}_{s^+\sim p_{\text {true }}(. \mid c)}\left[\log \mathcal{V}_{\phi}\left(c, s^{+}\right)\right]+\mathbb{E}_{s^{-} \sim p_{\text {neg }}\left(\cdot| \mathcal{G}_{\theta}, c, s^+\right)}\left[\log \left(1-\mathcal{V}_{\phi}\left(c, s^{-}\right)\right)\right].
$$
is reducible to the optimization problem of KL-divergence in Generator training in (Eq. 7).
First of all, we can discard the irrelevant terms of Eq. 2:

\begin{equation}
\begin{aligned}
J^{\mathcal{G}^{*}} &= \min _{\theta} \mathbb{E}_{s^{-} \sim p\left(\cdot| \mathcal{G}_{\theta}, c, s^+\right)}\log  \left(1-\mathcal{V}_{\phi}\left(c, s^{-}\right)\right) \\
&= \max _{\theta} \mathbb{E}_{s^{-} \sim p\left(\cdot| \mathcal{G}_{\theta}, c, s^+\right)}\log \left(\mathcal{V}_{\phi}\left(c, s^{-}\right)\right) \\
&\approx \max _{\theta} \mathbb{E}_{s^{-} \sim p\left(\cdot| \mathcal{G}_{\theta}, c, s^+\right)} \mathcal{V}_{\phi}\left(c, s^{-}\right).
\end{aligned}
\end{equation}
% 第三步可以说。。由于log的取值范围很大，在计算时很有可能越界，所以我们由第二步转化为第三步。（这个我不是很确定能不能这样整，如果从第二步推的话就没办法完美推出交叉熵了）
Since the sampling process of $s^-$ is discrete, we cannot directly optimize the $\mathcal{G}^*$ with gradient descent.
Following previous works in Sequential GAN, we apply the policy gradient approach.

\begin{equation}
\begin{aligned}
\nabla_{\theta} \hat{J}^{{G}^{*}} &=\nabla_{\theta} \mathbb{E}_{s^{-} \sim p_{\theta}\left(\cdot \mid c\right)} \mathcal{V}_{\phi}\left(c, s^{-}\right) \\
&=\sum_{i} \nabla_{\theta} p_{\theta}\left(s_{i}^{-} \mid c\right) \mathcal{V}_{\phi}\left(c,s_i^-\right) \\
&=\sum_{i}  p_{\theta}\left(s_{i}^{-} \mid c\right) \nabla_{\theta} \log p_{\theta}\left(s_{i}^{-} \mid c \right) \mathcal{V}_{\phi}\left(c,s_i^-\right)  \\
&= \mathbb{E}_{s^{-}} \left[  \nabla_{\theta} \log p_{\theta}\left(s_{i}^{-} \mid c \right) \mathcal{V}_{\phi}\left(c,s_i^-\right) \right]  \\
&\approx \frac{1}{K} \sum_{k=1}^K   \nabla_{\theta} \log p_{\theta}\left(s_{i}^{-} \mid c\right) \mathcal{V}_{\phi}\left(c,s_k^-\right)  \\
% &= \nabla_{\theta} \frac{1}{K} \sum_{k=1}^K    \log p_{\theta}\left(s_{i}^{-} \mid c,s^+\right) \mathcal{V}_{\phi}\left(c,s_k^-\right)  \\
% &= \nabla_{\theta} \text{CrossEntropy}(p_{\theta}\left(\cdot \mid c,s^+\right) \mathcal{V}_{\phi}\left(c,s_k^-\right) ) \\
&= \nabla_{\theta} \frac{1}{K} \sum_{k=1}^K \left[  -\mathcal{V}_{\phi}\left(c,s_i^-\right)\log \mathcal{V}_{\phi}\left(c,s_i^-\right) +\log p_{\theta}\left(s_{i}^{-} \mid c\right) \mathcal{V}_{\phi}\left(c,s_i^-\right) \right] \\
&= \nabla_{\theta}- D_{\text{KL}}(\mathcal{V}_{\phi}\left(c,s_k^-\right),p_{\theta}\left(\cdot \mid c \right)) \\
% &= \nabla_{\theta} \frac{1}{K} \sum_{k=1}^K \left[  \log p_{\theta}\left(s_{i}^{-} \mid c,s^+\right) \mathcal{V}_{\phi}\left(c,s_i^-\right) \right] \\
% &= \nabla_{\theta} \text{CrossEntropy}(p_{\theta}\left(\cdot \mid c,s^+\right) \mathcal{V}_{\phi}\left(c,s_i^-\right) )
\end{aligned}
\end{equation}
% 其实 KL散度的梯度和交叉熵是一样的，只是KL散度的公式前面有一个p_teacher*log P_teacher,这个项不贡献梯度。
which is equivalent to minimizing $\mathcal{L}_{gen}$ as decribed in Eq. 7 from the main text.
% the optimization of which is essentially equivalent with KL-divergence (Eq. 7 in the main text).

\end{document}